\documentclass[letterpaper, 10 pt, conference]{ieeeconf}  

\IEEEoverridecommandlockouts                              

\usepackage{graphics} 
\usepackage[caption=false, font=footnotesize,position=bottom]{subfig}
\usepackage{graphicx}
\usepackage{epsfig} 
\usepackage{amsmath} 
\usepackage{amssymb}  
\usepackage{gensymb}
\usepackage{booktabs}
\DeclareMathOperator*{\argmax}{argmax}

\title{\LARGE \bf
Cycle and Semantic Consistent Adversarial Domain Adaptation for Reducing Simulation-to-Real Domain Shift in LiDAR Bird’s Eye View
}

\author{Alejandro Barrera$^{1}$, Jorge Beltr\'an$^{1}$, Carlos Guindel$^{1}$, Jose Antonio Iglesias$^{2}$ and Fernando Garc\'i{}a$^{1}$%
\thanks{$^{1}$Authors are with the Dept. of Systems Engineering and Automation, Universidad Carlos III de Madrid, Spain {\tt\small alebarre@pa.uc3m.es, \{cguindel,jbeltran,fegarcia\}@ing.uc3m.es}} 
\thanks{$^{2}$J.A. Iglesias is with the Dept. of Computer Science and Engineering, Universidad Carlos III de Madrid, Spain {\tt\small jiglesia@inf.uc3m.es}}
}

\begin{document}

\maketitle
\thispagestyle{empty}
\pagestyle{empty}

\newcommand{\Lagr}{\mathcal{L}}
\newcommand{\E}{\mathbb{E}}
\newcommand{\norm}[1]{\| #1 \|}

\begin{abstract}
The performance of object detection methods based on LiDAR information is heavily impacted by the availability of training data, usually limited to certain laser devices. As a result, the use of synthetic data is becoming popular when training neural network models, as both sensor specifications and driving scenarios can be generated ad-hoc. However, bridging the gap between virtual and real environments is still an open challenge, as current simulators cannot completely mimic real LiDAR operation.
To tackle this issue, domain adaptation strategies are usually applied, obtaining remarkable results on vehicle detection when applied to range view (RV) and bird's eye view (BEV) projections while failing for smaller road agents. 
In this paper, we present a BEV domain adaptation method based on CycleGAN that uses prior semantic classification in order to preserve the information of small objects of interest during the domain adaptation process. The quality of the generated BEVs has been evaluated using a state-of-the-art 3D object detection framework at KITTI 3D Object Detection Benchmark. The obtained results show the advantages of the proposed method over the existing alternatives.
\end{abstract}
\section{INTRODUCTION} \label{intro}

Nowadays, perception is a crucial task for autonomous vehicles. Research in this field demands accurate sensors and 
algorithms to perform safe and precise navigation. LiDAR stands as an ideal candidate to directly describe the scene geometry by a dense point cloud representation.


Despite the recent increment in the amount of labeled data, public datasets may not be sufficient to train models able to grasp a complete understanding of the situations that they may encounter in operation due to the domain shift problem. Variations in sensor positions, device specifications (e.g., number and distribution of planes), or even the geographic region \cite{train_germ_usa} can lead to a significant performance drop in supervised learning approaches. Moreover, the annotation type (point-wise, 3D boxes, etc.) could also differ between source and target samples, and collecting well-annotated data for custom applications is prohibitively expensive. 

Hence, synthetic data stands as an enticing option to provide on-demand and accurate data which can be modified and extended almost infinitely. Despite the realism of current simulators' sensor and world models available today, algorithms trained with these data usually fail to generalize in a real environment.

Domain adaptation (DA) techniques have been explored to bridge the aforementioned gaps between domains. Therefore, some approaches have attempted to directly adapt raw LiDAR information to other data distributions \cite{pointgan1}. Nevertheless, due to the sparsity, irregularity, and unstructured distribution of LiDAR data and the high number of points contained in each cloud, on-board perception applications often use efficient projections such as the range view (RV) and the bird's eye view (BEV), for which DA alternatives have also been proposed \cite{epointda, bevda1}.

In many of these works, CycleGAN \cite{cyclegan} and its cycle consistency mechanism have reported an excellent performance on image-level domain adaptation and content preservation for these 2D projected-based representations. Whilst this method can produce realistic adaptations of big and medium-sized vehicles, we argue that further refinement \cite{cycada, semgan} can help preserve scarcely represented objects, which are normally vulnerable road users such as pedestrians and cyclists.



\begin{figure}[thb]
\centering
\includegraphics[width=\linewidth]{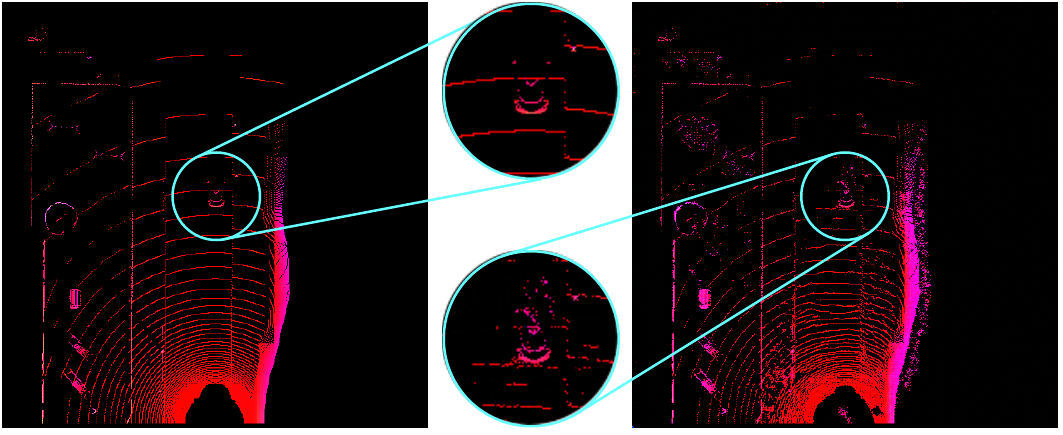}
\caption{On the left, raw synthetic BEV recorded in the CARLA simulator. On the right, the result of the proposed domain adaptation. Zoomed regions are provided to better observe the differences.}
\label{fig:init_diagram}
\end{figure}

This work proposes an approach to enhance the style transfer of BEV representations from a synthetic scalable source domain, generated using a simulator, to a real target domain. 
The conversion, shown in Fig.~\ref{fig:init_diagram}, makes use of an adversarial generative network adapted to BEV representations and endowed with semantic segmentation consistency to help preserve object instances in the scene. 
To the best of our knowledge, this is the first method addressing unsupervised domain adaptation between unpaired BEV images using a CycleGAN with multi-class semantic regularization.

The goal is to avoid, or at least reduce, the need for labeled samples from the target domain, thus enabling the deployment of high-complexity models on custom setups. In particular, to validate the effectiveness of the approach, we use adapted synthetic data to train a state-of-the-art BEV-based 3D object detection method, BirdNet+ \cite{birdnet+}, which is later deployed and tested on the KITTI object detection benchmark \cite{kitti}.




\section{Related Work} \label{soa}

The advent of deep learning has led to a significant research interest in domain adaptation (DA).
Among deep DA methods, adversarial-based ones generally use a domain classifier to extract domain-invariant features through the confusion of the source and target domain boundaries \cite{dagans}. The optimum result from these networks is to minimize the domain distance to maximize the domain discriminator error, producing data that the discriminator cannot distinguish from real \cite{gans, improvedgans}. On the other hand, in the reconstruction-based DA category, \cite{johnsonstyle} combines different losses to recombine style and content from two separate images.


CycleGAN \cite{cyclegan} combines both solutions using an adversarial loss and a reconstruction loss (cycle consistency loss) to address the image-to-image translation problem when paired training data is not available (unsupervised DA or UDA). 
Although the CycleGAN reconstruction task shows promising results in a wide variety of scenarios, CYCADA \cite{cycada} extends its capabilities using both image space alignment and latent representation space alignment. Besides, it incorporates a task to encourage content consistency enforcing relevant semantics to match before and after adaptation.
This semantic consistency has proven vital in multimodal scenarios because the invertibility provided by the cycle does not necessarily preserve the arrangement of the classes from the original source domain, as shown in \cite{semgan}. However, unlike our proposal, this method requires labels of both the source and target domains, as it operates in a supervised fashion.

All 
these methods are designed for 2D vision tasks where RGB images are the protagonists. However, when it comes to LiDAR point cloud representations, 
some adaptations are required.
In order to work with point clouds, the most straightforward alternative to preserve all the LiDAR information is to use raw clouds to perform point-wise DA and set-level DA \cite{pointgan1}. By the same token, PointDAN \cite{pointdan} studies local-level and global-level point cloud alignments by the use of self-adaptive attention nodes. 

Although such methods are able to preserve all the LiDAR information, their execution time and memory requirements make them inefficient when it comes to a full point cloud. In this context, projection-based methods gain popularity due to their adaptability to the well-studied 2D approaches. Unfortunately, this also entails the inevitable loss of spatial information. 
In this field, ePointDA \cite{epointda} uses range 
view representations from simulation and real domains to bridge the domain gap at pixel-level and feature-level. 
LiDARNet \cite{lidarnet} combines multiple tasks such as boundary extraction, cycle consistency, and domain invariance to address a full-scene semantic segmentation task on real range view images. BEV-Seg \cite{bev-seg} uses multiple camera angles, with RGB and depth images from a simulator to create a semantically enriched point cloud to find BEV semantic segmentation.

Regarding BEV projection, \cite{bevda1} generates from simulation data realistic scenarios to transfer annotations from each other, and \cite{bevda} shows the capabilities of a similar method on a BEV-based detector. 

As can be seen, many of the previous works focus their efforts on simulation-to-real domain adaptation (SRDA). The main reason is to avoid the very challenging annotation task, which is a time and money-consuming task. Considering this issue, simulators such as CARLA \cite{carla}, which counts with multiple modeled sensors, or LiDAR-based datasets such as GTA-V \cite{gtav} and SynthCity \cite{synthcity} have been developed. 
Furthermore, some works improve existing synthetic data adding well-modeled obstacles where needed \cite{syntheticlidar}. 





\section{Proposed Method} \label{method_head}
\begin{figure*}[tb]
\centering
\includegraphics[width=0.9\linewidth]{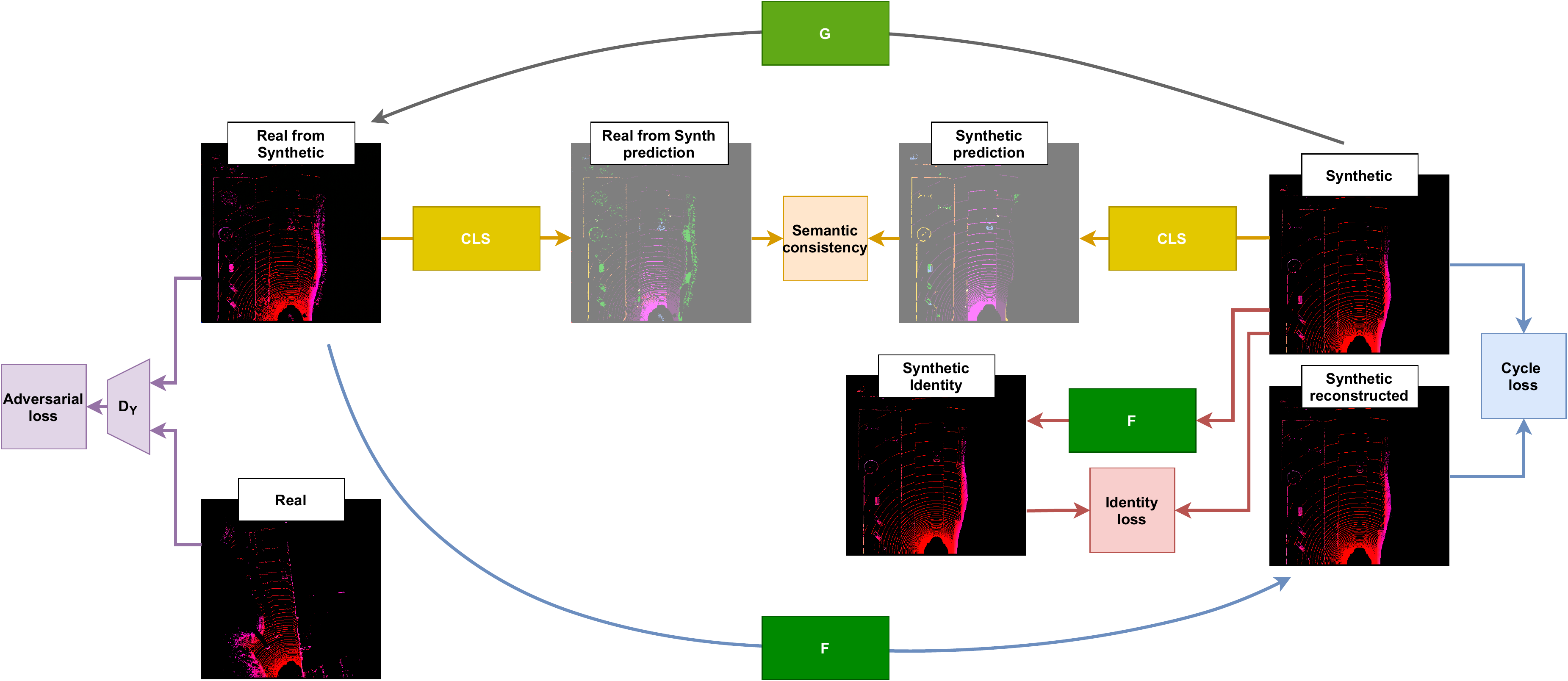}
\caption{Overview of the proposed approach. We depict the adversarial loss, to transform the domains, in purple, the cycle consistency mechanism in blue, the identity consistency in red, and the semantic constraint in orange. 
Blocks $G$ and $F$ stand for the generators $G: X \rightarrow Y$ and $F: Y \rightarrow X$, $D_Y$ for the discriminator of the real domain, and CLS for the semantic segmentation network. For clarity reasons, only losses where the synthetic representation is involved are shown. The real cycle is designed to mirror the synthetic one.}
\label{fig:network_diagram}
\end{figure*}

This section provides a detailed explanation of the proposed approach, which is depicted in Fig.~\ref{fig:network_diagram}. Two different sets of bird's eye view (BEV) images, encoding LiDAR information, are used as input to carry out a transformation between unpaired synthetic BEV point cloud data and real data.
This fact will make it possible to use annotations from synthetic data in place of real data, and therefore, expand and improve the diversity of the objects for the detection task.


\subsection{Input Representation} \label{input_repr}
Our BEV representation follows the one proposed in \cite{birdnet+}; however, we dispense with the LiDAR intensity data, for which realistic values are difficult to obtain. Thus, three distance-invariant channels are used: maximum height, normalized density within each cell, and binary occupancy. 
In our experiments, we use a voxel size of $10$~cm to obtain handleable representations and a data range of $50$~m forward and $22.5$~m to each side in order to represent the area where the majority of annotations are available in the KITTI dataset.

For the generation of the synthetic BEV images, we rely on the CARLA simulator \cite{carla}. This simulator provides multiple realistic scenarios, agent models, and sensors generated by the graphics engine Unreal.
We use a semantically enhanced LiDAR 
modeled after the KITTI dataset device and let the domain adaptation model the noise. 


\subsection{Architecture Description} \label{arch_desc}


Our adversarial-based approach, which provides a translation of both representations guided by cycle, identity, and strong pixel-level semantic-aware consistencies, is built upon the CycleGAN architecture \cite{cyclegan}.

\medskip
\noindent \textbf{Adversarial network.} 
Given the source domain $X$ (synthetic BEV images) and the target domain $Y$ (real data), an adversarial network aims to map the data distributions of each domain, $x \sim p_d(x)$ and $y \sim p_d(y)$, to the other. 
First, the architecture is composed of two classifiers named domain discriminators, $D_X$ and $D_Y$, that learn the characteristic features of each domain separately. Then a set of generators $G$ and $F$ is designed to translate $G: X \rightarrow Y$ and $F: Y \rightarrow X$. Finally, the discriminators $D_X$ and $D_Y$ will provide the necessary feedback of the undergoing mapping $G(x)$ and $F(y)$ until they can not distinguish the domains. 

Generators $G$ and $F$, following the architecture in \cite{johnsonstyle}, are organized as follows: first, the encoder module reduces the initial resolution smoothly by a downsampling factor of~$4$. 
Afterward, the transformer, which leads the conversion between domains, is built with nine residual blocks that follow a structure similar to that of the encoder, 
but to which strong dropout is applied during training to provide noise \cite{patchgan_pix2pix}.
Finally, the decoder mimics the encoder structure, but it contains fractionally strided convolutions and one convolution to map features to RGB values, followed by a $\tanh$ activation function. 

On the other hand, discriminator networks are inspired in the PatchGAN architecture \cite{patchgan_pix2pix}, where four sets of convolution + instance normalization + LeakyReLU layers downsample the initial resolution and then, two non-strided convolutions predict on $70 \times 70$-pixel overlapping image patches the domain to which they belong.

\medskip
\noindent \textbf{Cycle consistency.} Although adversarial training produces some resemblance between each domain results, it still lacks a mechanism to preserve the structure from the original domain after the conversion. Therefore, in \cite{cyclegan}, a cyclical training is proposed to prevent $G$ and $F$ from contradicting each other; thus, the mapping $G( F (y) )$ will attempt to reproduce the content in $y$ and $F( G (x) )$ the content in $x$.

\medskip
\noindent \textbf{Identity consistency.} One more constrain is imposed on the generators to minimize their distance to an identity mapping when real samples of the opposite domain are provided: $y \approx G(y)$ and $x \approx F(x)$. This idea generally preserves better the information contained in each channel, which may otherwise suffer from undesirable blending under adversarial training \cite{cyclegan}. 
Working with a BEV representation requires retaining the per-pixel structure of the channels so that crucial information, such as object height or cell density, is not modified in the process. 


\medskip
\noindent \textbf{Semantic consistency.} 
Unlike \cite{bevda}, Sem-GAN \cite{semgan} demonstrates that the previous methods (i.e., cycle and identity consistency) do not necessarily maintain object identities, but they focus on the whole image adaptation instead. It implies that small details in a BEV representation such as poles and pedestrians could disappear during the transformation.
In order to encourage source voxels to keep their structure while being translated to the target domain, we follow the approach in CYCADA~\cite{cycada}. The idea consists of training a semantic segmentation network on the synthetic domain beforehand and using its predictions to ensure high semantic consistency after conversion, thus preserving the fine-grained content and the style of the input.

As our method is unsupervised, we only use synthetic labels in the process. In addition, for the semantic segmentation task, we chose a network that has previously provided state-of-the-art results in LiDAR projection-based representations such as SalsaNext \cite{salsanext}. Although not dedicated to operating in BEV representations, its predecessor was able to use both representations indistinctly.
This network, arranged in an encoder-decoder fashion, is composed of a contextual module that stacks three residual blocks to fuse features of two different receptive fields. Afterward, the network follows a U-Net-based structure concatenating residual blocks $i$ with the $n - i$ blocks. 
Similarly, the decoder utilizes a sequence that mimics the encoder; however, it is preceded by a pixel-shuffle layer. 



\subsection{Loss Functions} 
\label{loss_head}
To train the proposed adversarial network, two different cost functions are minimized. On the one hand, each discriminator $D$ tries to reduce, independently of the generator, an adversarial loss $\Lagr_{\text{adv}_D}$; hence, for $D_Y$: 
\begin{equation}
  \begin{split}
    \Lagr_{\text{adv}_{D_Y}} &= \E_{y \sim p_d(y)} [(D_Y(y) - 1 )^2] + \\ &\hphantom{{}={}} \E_{x \sim p_d(x)} [ D_Y( G(x) )^2 ]
  \end{split}
\label{eq:advloss_D_XY}
\end{equation}
In this way, the classifier is trained to distinguish between its domain ($D_Y(y) \approx 1$) and the domain representation created by the opposite generator ($D_Y(G(x)) \approx 0$).

On the other hand, the generators' multi-task training loss is given by the following equation, where each component has a weight $\lambda$ and is computed in both directions; i.e., $X\rightarrow Y$ and $Y\rightarrow X$:
\begin{equation}
    \Lagr = \Lagr_{\text{adv}_\text{gen}} +  \lambda_\text{cyc}\Lagr_\text{cyc} + \lambda_\text{idt}\Lagr_\text{idt} + \lambda_\text{sem}\Lagr_\text{sem}
    \label{eq:totalloss}
\end{equation}

The generators attempt to minimize the adversarial loss $\Lagr_{\text{adv}_\text{gen}}$, where the usual negative log-likelihood or binary cross-entropy loss has been modified by a least-squares loss, as in \cite{cyclegan}:
\begin{equation}
\begin{split}
    \Lagr_{\text{adv}_\text{gen}} &= \E_{x \sim p_d(x)} [( D_Y( G(x) ) - 1 )^2 ] + \\ &\hphantom{{}={}} \E_{x \sim p_d(y)} [( D_X( F(y) ) - 1 )^2 ] 
\end{split}
\label{eq:advloss_G_XY}
\end{equation}

The first reconstruction component, namely cycle consistency $\Lagr_{cyc}$, is an L1 penalty between the initial sample from one domain to the final representation, after both translations $x \rightarrow G( x ) \rightarrow F( G( x ) ) \approx x$ (and the analogous for domain $Y$):
\begin{equation}
  \begin{split}
    \Lagr_\text{cyc} &= \E_{x \sim p_d(x)} [\norm{F(G(x)) - x}_1] + \\ &\hphantom{{}={}} \E_{y \sim p_d(y)} [\norm{G(F(y)) - y}_1]
  \end{split}
\label{eq:cycloss}
\end{equation}

Secondly, the identity loss is defined as the mean absolute error, L1 loss, to ensure when a generator is fed by a sample from the opposite domain, it can produce a representation of its own domain:
\begin{equation}
  \begin{split}
    \Lagr_\text{idt} &= \E_{x \sim p_d(x)} [\norm{G(y) - y}_1] + \\ &\hphantom{{}={}} \E_{y \sim p_d(y)} [\norm{F(x) - x}_1]
  \end{split}
\label{eq:idtloss}
\end{equation}

Finally, regarding the semantic loss, we use the SalsaNext \cite{salsanext} semantic segmentation outputs as a noisy labeler to keep as much context as possible after translation:
\begin{equation} \label{eq:loss_G_sem}
\begin{split}
\Lagr_\text{sem} &= \Lagr_\text{wCE}(\text{CLS}(G(x)), \argmax{(\text{CLS}(x))} ) + \\
            &\hphantom{{}={}} \Lagr_\text{wCE}(\text{CLS}(F(y)), \argmax{(\text{CLS}(y))} )
\end{split}
\end{equation}
where $\Lagr_\text{wCE}$ represents a weighted cross-entropy loss, which is computed over the pixels of the semantic prediction ($\text{CLS}$) for $G(x)$ and $F(y)$ with the predictions from the source BEVs ($x$ and $y$, respectively) as weak labels. Only cells with non-zero values in both inputs contribute to the loss in order to preserve both the class and location of the points. Weights are used to increase the importance, by a $2\times$ factor, of the categories of interest (i.e., cars, pedestrians, and cyclists) over the rest of classes (e.g., roads or buildings), which are still included to maintain the geometric consistency of the complete scene.

As stated above, we use a SalsaNext model as semantic predictor ($\text{CLS}$). This model is trained beforehand with synthetic data through the usual weighted multi-category cross-entropy and Lovász-Softmax losses.

\section{Experimental Results} \label{experiments_head}
In this section, we present a set of experiments to validate the performance of our domain adaptation approach from synthetic BEV data to real BEV representations. Our implementation will be evaluated on the well-known KITTI object dataset using the 3D detector specified below.

\subsection{3D Obstacle Detection for Evaluation} \label{det_3d}
Assessing the feasibility of a domain adaptation method on a 3D detector is a common practice nowadays. The main advantage lies in the fact that the results will offer a good estimate of the resemblance of the generated data to the real dataset to be emulated. 
With this task in mind, we have chosen a 3D detector that uses enriched BEV inputs to provide the object's location, shape, and category in a two-stage end-to-end fashion. The first stage of the BirdNet+ architecture \cite{birdnet+} is built upon a residual-based encoder with per-level skips to preserve global and local content (ResNet-50 and a feature pyramid network). These features are fed into a region proposal network that classifies and refines a default anchor estimation. These non-axis-aligned proposals are dimensionally normalized by an ROI Align layer and forwarded to a second stage composed of a sequence of multiple fully-connected layers, which finally estimate the 3D object parameters.

\subsection{Experimental Setup} \label{train_dets}
As in the original CycleGAN approach \cite{cyclegan}, the adversarial network is trained from scratch with random weights following a normal distribution 
$\mathcal{N}(0,\,0.02)$ 
to initialize the weights of every layer in our model. Training data is randomly augmented using different techniques: horizontal flip, point dropout, and additive Gaussian noise with similar distribution to our input representation. Following \cite{improvedgans}, real labels in (\ref{eq:advloss_D_XY}) and (\ref{eq:advloss_G_XY}) are softened randomly and kept between $0.7$ and $1$. Additionally, the last $50$ samples are used to compute the discriminators' losses and provide better stability.

For our experiments, we fix $\lambda_\text{cyc} = \lambda_\text{idt} = 10$ and $\lambda_\text{sem} = 0.5$ in (\ref{eq:totalloss}) to weigh all the losses. We use the Adam solver for the optimization with momentum $[0.5, 0.99]$ and train up to $50$ epochs. The number of epochs, batch size, learning rate (LR), and learning rate decay of all networks involved in this work are indicated in Table~\ref{tab:train_params}.

\begin{table}[htbp]
\caption{Training Parameters for the Different Models.}
\label{tab:train_params}
\begin{center}
\begin{tabular}{l r c l l}
\toprule
\textbf{Network} & \textbf{Epochs} & \textbf{Batch} & \textbf{LR} & \textbf{LR decay}\\
\midrule
DA network  & 50 & 1 & 0.0001 & None\\
SalsaNext & 100 & 1 & 0.01 & 0.01 every epoch\\
BirdNet+  & $\sim$12 & 4 & 0.01 & 0.1 at the 10\textsuperscript{th} epoch\\
\bottomrule
\end{tabular}
\end{center}
\end{table}

Our synthetic data have been extracted from five different towns 
delivered by the CARLA simulator. Between $1000$ to $1500$ samples per town were extracted with a delay of $0.5$ simulated seconds to provide more variability to our results. It is worth noting that in one of the maps, we only use cyclists and pedestrians to deal with the unbalanced data. 

The scenes contain a limited and approximately constant number of agents, which are spawned and destroyed outside the field of view of our BEV representation to avoid inconsistencies. 
Besides, we divide the CARLA vehicle class into more fine-grained categories (i.e., car, motorcycle, and bicycle), matching the KITTI labeling criteria for two-wheelers (so that they include both the vehicle and the rider).
Additionally, 3D labels for parked vehicles, which are not reported by CARLA, are estimated using the dense BEV semantic representation provided by a top-view camera.


In total, $6878$ synthetic BEV images, endowed with semantic and 3D box labels, are extracted. They are used to train both SalsaNext and the proposed domain adaptation framework (together with the KITTI training set as target). Finally, the 3D detection model is trained with the adapted images (considering only the car, pedestrian, and cyclist categories) and tested on the KITTI validation set, defined as in \cite{birdnet+}.




\subsection{Overall assessment} \label{results_head}
\begin{figure}
    \subfloat{%
    \includegraphics[width=0.31\linewidth]{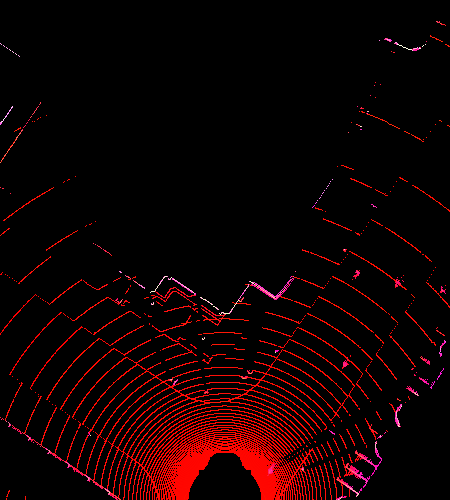}} \hfill
   \subfloat{%
    \includegraphics[width=0.31\linewidth]{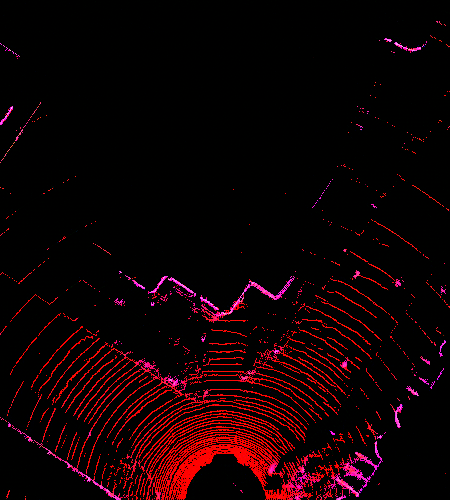}} \hfill
   \subfloat{%
    \includegraphics[width=0.31\linewidth]{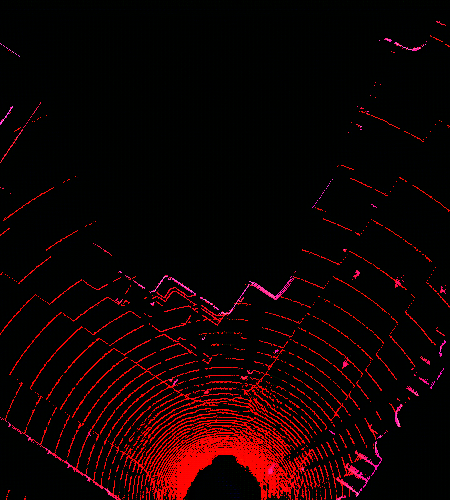}} \setcounter{subfigure}{0} \\ 
   \subfloat[Synthetic]{%
   \includegraphics[width=0.31\linewidth]{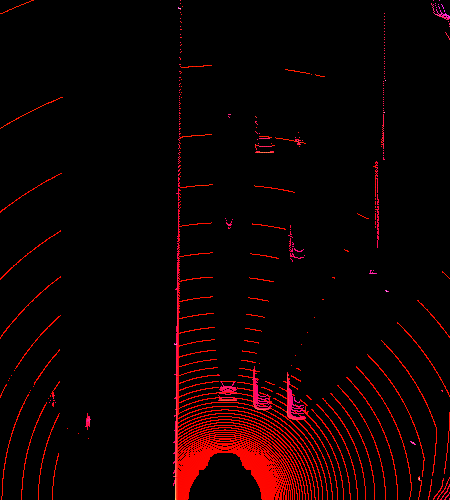}\label{fig:1s}} \hfill 
   \subfloat[Original DA]{%
    \includegraphics[width=0.31\linewidth]{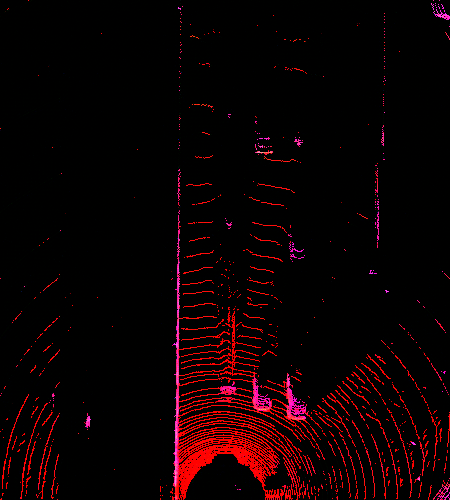}\label{fig:2da}} \hfill 
   \subfloat[Ours]{%
    \includegraphics[width=0.31\linewidth]{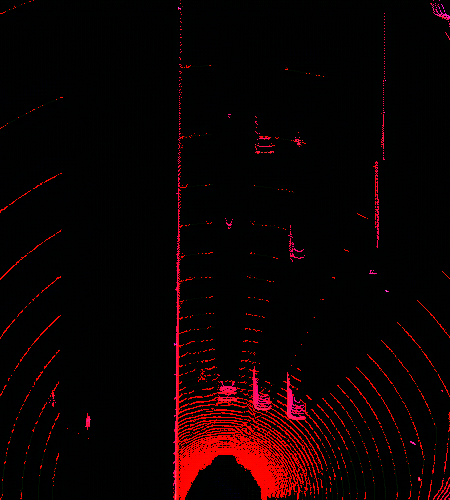}\label{fig:3das}} \hfill 
\caption{Domain adaptation results with the original DA (CycleGAN) and the proposed approach.} \label{fig:da_comparison}
\end{figure}

To shed light on the importance of the proposed consistency in the domain adaptation, Fig.~\ref{fig:da_comparison} shows a sample from the CARLA simulator, the output after the adversarial training proposed in~\cite{cyclegan}, and the output with our model, including all the losses described in Sec.~\ref{loss_head}.

The noise introduced in the original DA approach (Fig.~\ref{fig:2da}) adds unrealistic LiDAR measurements and changes the pixel value of some areas. On the other hand, our method (Fig.~\ref{fig:3das}) seems to eliminate some points from the ground that may impact the segmentation task, but it preserves significantly better the semantic identity of each individual pixel while modeling the noise of the real LiDAR.


For the evaluation of the performance of the BirdNet+ detector trained with the adapted data, we follow the 3D and BEV detection tasks from the KITTI object detection benchmark \cite{kitti}.
The strong IoU requirements between detections and labels imposed by the official evaluation do not fit well the localization uncertainty shown by models trained only with synthetic data; therefore, we employ less strict thresholds: $50\%$ (cars) and $30\%$ (pedestrians and cyclists).
Table \ref{tab:bev3d_da_comparison} shows the considerable performance gap in the domain shift between the synthetic and the real in our representations. It is worth noting that the original DA method focuses either on medium and big-sized obstacles (i.e., cars) and structures rather than small objects, which, in the end, are partially occluded by the noise generated. In our approach, the disappearance or modification of these elements penalizes the generator, that becomes more aware of the semantic context of each point. Thus, our method preserves better the details in the scene, easing the task of 3D object detection in BEV representations. 

\begin{table}[htb] 
	\caption{BirdNet+ Detection Performance (AP BEV \% and AP 3D \%) on the KITTI Val Set for the Different Training Data Sources.}
	\label{tab:bev3d_da_comparison}
	\centering
	\begin{tabular}{l c c c c c c }
		\toprule
		& \multicolumn{2}{c}{Car} & \multicolumn{2}{c}{Pedestrian} & \multicolumn{2}{c}{Cyclist} \\  
		\cmidrule(lr){2-3} \cmidrule(l){4-5} \cmidrule(l){6-7}
		 & BEV & 3D & BEV & 3D & BEV & 3D \\ 
		 \midrule   
		 Oracle (KITTI) & 81.94 & 67.04 & 50.17 & 43.90 & 42.74 & 39.89 \\ 
		 \midrule
         Synthetic & 52.91 & 46.82 & 18.11 & 17.91 & 22.37 & 21.85  \\ 
         Original DA & \textbf{61.53} & 48.08 & 10.58 & 07.45 & 18.82 & 16.56 \\
         Ours & 53.79 & \textbf{48.61} & \textbf{26.21} & \textbf{25.81} & \textbf{29.88} & \textbf{29.75} \\
		\bottomrule
	\end{tabular}
\end{table}

\begin{figure*}[thb]
\captionsetup[subfloat]{farskip=4pt}
   \def\twidth{0.325}
   \subfloat{%
    \includegraphics[width=\twidth\linewidth]{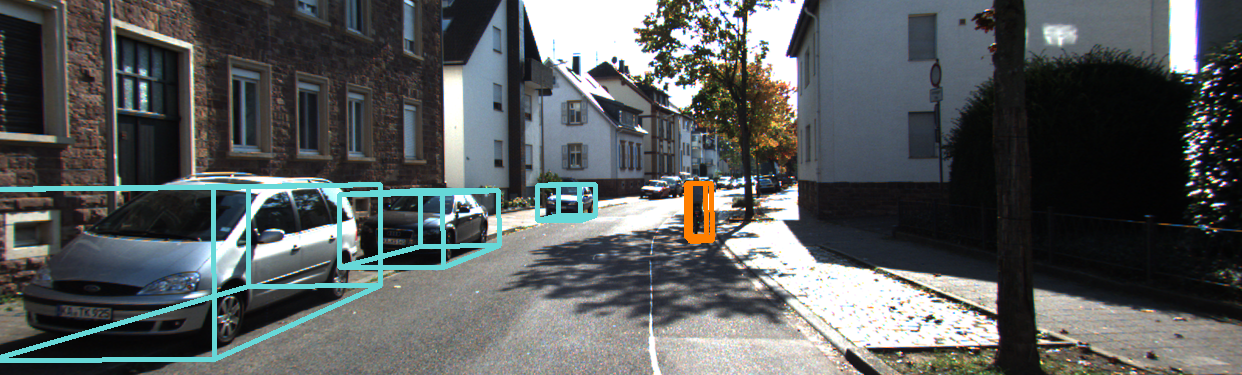}%
    }\hfill
    \setcounter{subfigure}{0}%
   \subfloat{%
    \includegraphics[width=\twidth\linewidth]{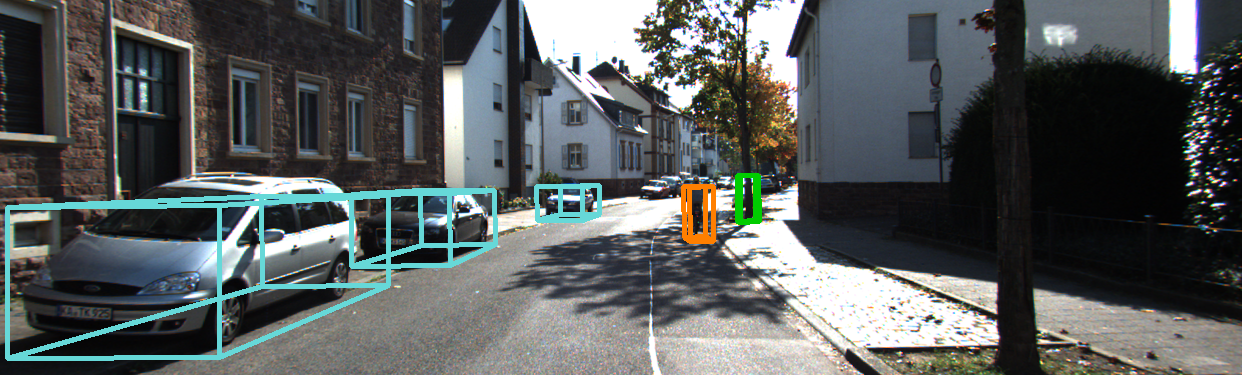}\label{fig:7232_qual}%
    }\hfill
   \subfloat{%
    \includegraphics[width=\twidth\linewidth]{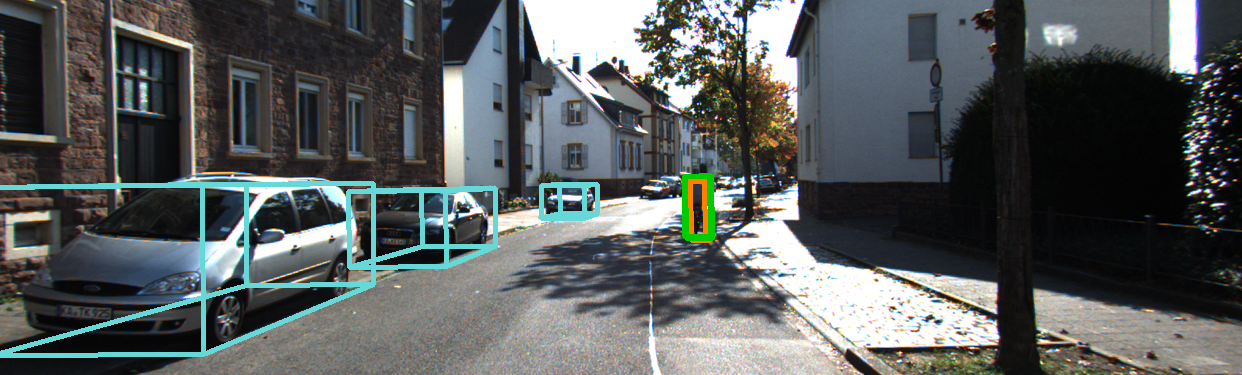}%
    }\\
   \subfloat{%
    \includegraphics[width=\twidth\linewidth]{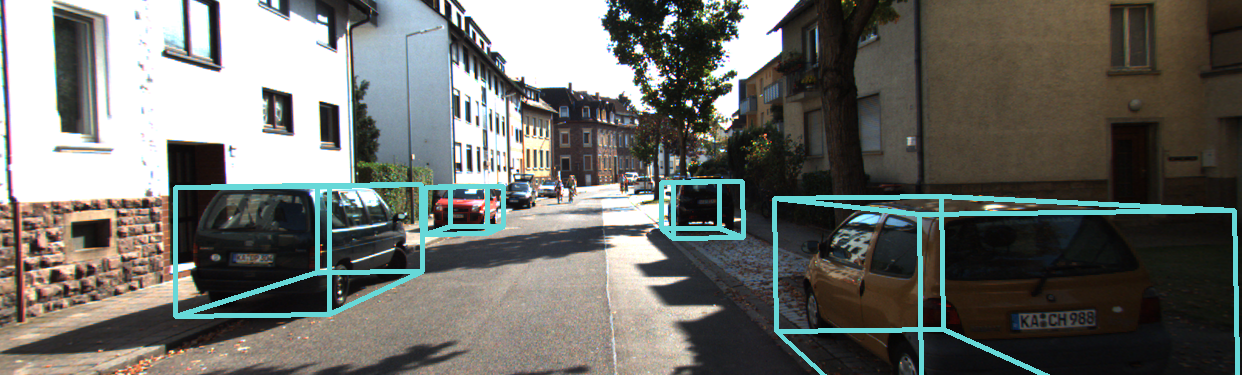}%
    }\hfill
    \setcounter{subfigure}{1}%
   \subfloat{%
    \includegraphics[width=\twidth\linewidth]{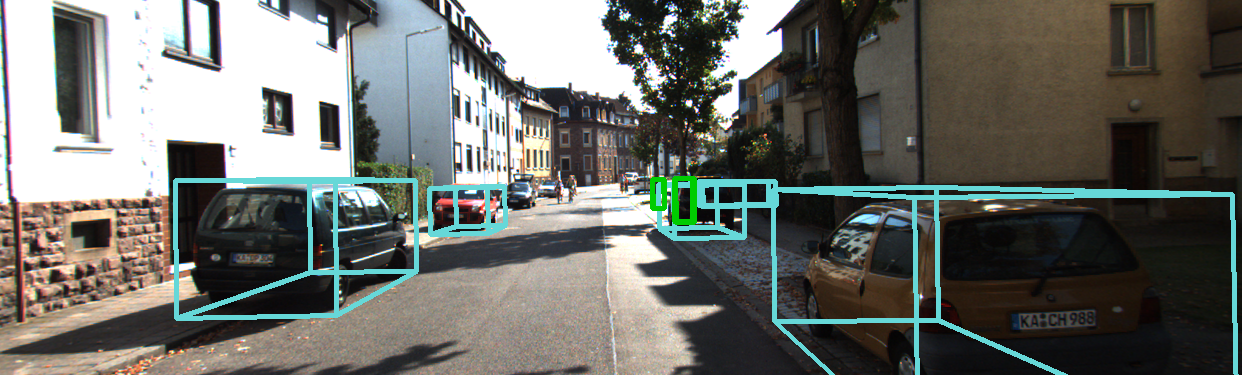}\label{fig:7277_qual}%
    }\hfill
   \subfloat{%
    \includegraphics[width=\twidth\linewidth]{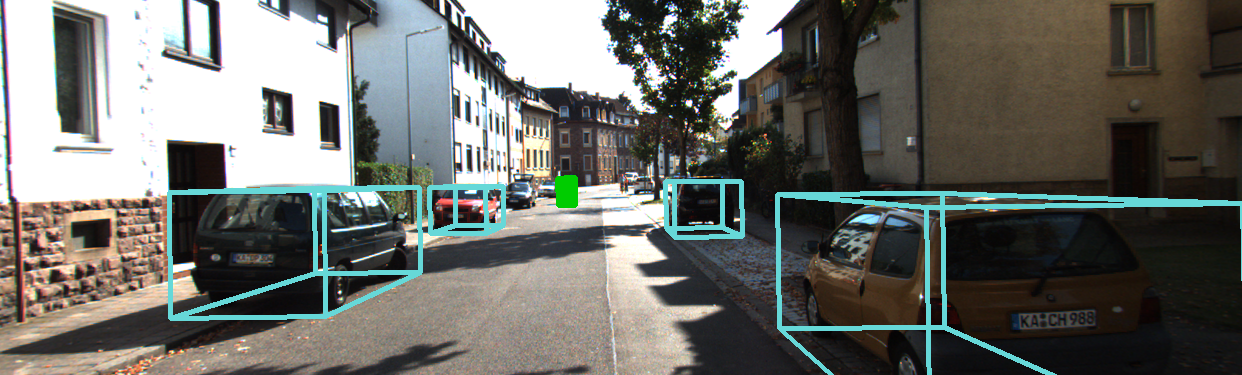}%
    }\\
   \subfloat{%
    \includegraphics[width=\twidth\linewidth]{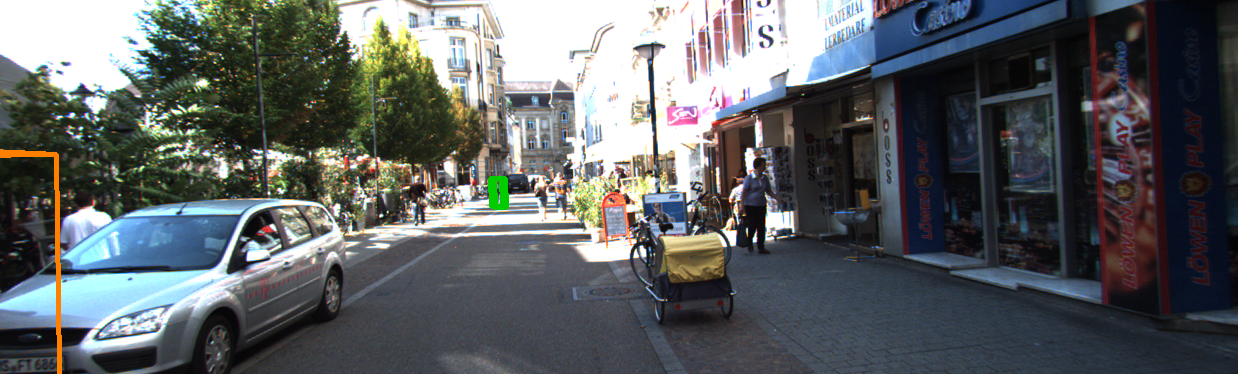}%
    }\hfill
    \setcounter{subfigure}{2}%
   \subfloat{%
    \includegraphics[width=\twidth\linewidth]{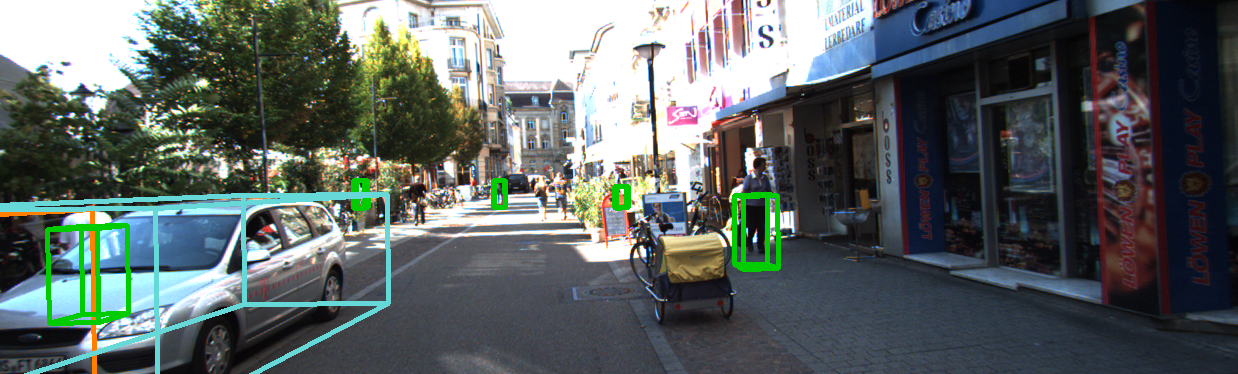}\label{fig:7464_qual}%
    }\hfill
   \subfloat{%
    \includegraphics[width=\twidth\linewidth]{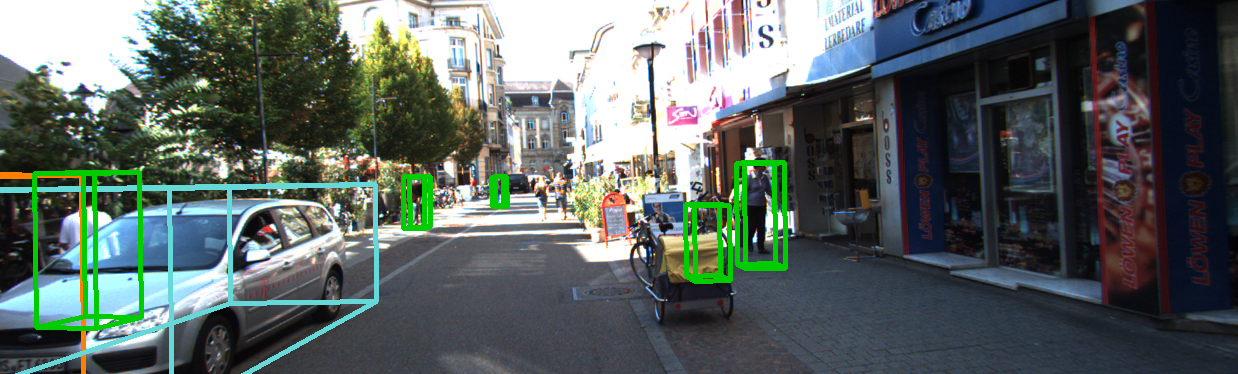}%
    }
\caption{
Qualitative results in KITTI validation set produced by BirdNet+ using different training data. From left to right: raw synthetic data, cycle consistent DA, cycle and semantic consistent DA.} \label{fig:kitti_qual}
\end{figure*}

The validity of the method is further confirmed in the qualitative results depicted in Fig.~\ref{fig:kitti_qual}.
As can be seen, our method adjusts the elevation of cars (first row) and pedestrians (third row) in a better way due to the fact that it preserves the pixel values better than the others whilst generating fewer artifacts. In addition, our method provides more recall in small classes (second and third rows) such as pedestrians and cyclists; however, it occasionally fails to distinguish between them. It is clear that, although detection capabilities are naturally limited by the domain gap, our method demonstrates a significant improvement over its predecessor and the synthetic-only approach in the 3D detection task.





\section{CONCLUSIONS}

In this work, it has been shown that the enforcement of semantic consistency in GAN-based domain adaptation of BEV projections benefits the preservation of the original layout of the elements in the synthetic scene during style transfer. The performance of the presented framework has been assessed using a state-of-the-art object detection network in the challenging KITTI Benchmark. Contrary to the baseline method, our results improve by a wide margin those obtained when training with raw synthetic data, being especially significant the difference in the detection of smaller road participants.

In future works, a lossless LiDAR style transfer will be studied so that any kind of object detection network can be used regardless of its input representation. To this aim, two different approaches will be explored: first, by means of several simultaneous generators per domain dedicated to different projections, which will ultimately allow reconstructing the LiDAR point cloud; and second, by designing a method able to perform domain adaptation over the raw 3D information.






\section*{ACKNOWLEDGMENT}

Research conducted within the project PEAVAUTO-CM-UC3M. The research project  PEAVAUTO-CM-UC3M has been funded by the call “Programa de apoyo a la realización de proyectos interdisciplinares de I+D para jóvenes investigadores de la Universidad Carlos III de Madrid 2019-2020 under the frame of the Convenio Plurianual Comunidad de Madrid-Universidad Carlos III de Madrid.
We gratefully acknowledge the support of NVIDIA Corporation with the donation of the GPUs used for this research.


\bibliographystyle{IEEEtran}
\bibliography{paper}

\end{document}